\documentclass[11pt]{article}

\usepackage[preprint]{acl}

\usepackage{amsmath,amsfonts,bm}

\def\eqref#1{equation~\ref{#1}}

\def\1{\bm{1}}

\DeclareMathAlphabet{\mathsfit}{\encodingdefault}{\sfdefault}{m}{sl}
\SetMathAlphabet{\mathsfit}{bold}{\encodingdefault}{\sfdefault}{bx}{n}

\DeclareMathOperator*{\argmin}{arg\,min}

\usepackage{times}
\usepackage{latexsym}
\usepackage{amsmath}
\usepackage{rotating}
\usepackage{multirow}
\usepackage{booktabs}
\usepackage{amsfonts}
\usepackage{subfig}

\usepackage[T1]{fontenc}
\usepackage{wrapfig,comment}

\usepackage[utf8]{inputenc}

\usepackage{microtype}

\usepackage{inconsolata}

\usepackage{graphicx}

\title{SLIM-LLMs: Modeling of Style-Sensory Language Relationships Through Low-Dimensional Representations
}

\author{Osama Khalid \\
  \texttt{} \\\And
  Sanvesh Srivastava \\
  \texttt{} \\\And
    Padmini Srinivasan \\
  \texttt{}}

\begin{document}
\maketitle
\begin{abstract}

Sensorial language --- the language connected to our senses including vision, sound, touch, taste, smell, and interoception --- plays a fundamental role in how we communicate experiences and perceptions. We explore the relationship between sensorial language and traditional stylistic features, like those measured by LIWC, using a novel Reduced-Rank Ridge Regression (R4) approach. We demonstrate that low-dimensional latent representations of LIWC features ($r = 24$) effectively capture stylistic information for sensorial language prediction compared to the full feature set ($r = 74$). We introduce Stylometrically Lean Interpretable Models (SLIM-LLMs), which model non-linear relationships between these style dimensions. Evaluated across five genres, SLIM-LLMs with low-rank LIWC features match the performance of full-scale language models while reducing parameters by up to $80\%$.
\end{abstract}

\section{Introduction}
Linguistic style includes traditional stylistic features like sentence length, language complexity, sentiment, and syntactic structure. It also includes patterns in the language used to describe sensory experiences --- sensorial style, a phenomenon that has only recently received attention in the stylometrics literature. 

Consider that a person might describe her feelings by using the word `sad' or she could use the more complex word `melancholic' to describe the same feelings. If she uses complex language frequently and consistently, it may be considered a part of her linguistic style. 
The same person having a cup of coffee, might focus on the taste of the coffee and describe it as `bitter' or she might instead focus on its warmth and can describe it as `hot'. This emphasis on one sensory modality over another would reflect her sensorial style. Alternatively, she might engage with multiple sensory aspects equally, describing both the coffee's temperature and its taste - this balanced sensory attention would also characterize her sensorial style.

Sensorial linguistics investigates the relationship between sensory perception and language, studying how different experiences and perceptions are represented using linguistic units \citep{bodosensory}. Sensorial style is a relatively new area of research and is informed by ideas from sensorial linguistics and holds significant potential for providing insights into human cognition. 

Standard stylometric lexicons, such as LIWC \cite{pennebaker2007linguistic}, do include some sensorial terms. However since the primary focus of LIWC is on psychological and cognitive aspects, the sensorial terms are generally distributed across different LIWC subcategories and LIWC's coverage of the sensorial language space is sparse\footnote{The LIWC 2015 \cite{pennebaker2015development} lexicon only covers 29\% of the sensorial vocabulary, defined by \citeauthor{khalid2022smells}}. 
The same is true for other lexicons like ANEW \citep{bradley1999affective}. 

 Understanding how traditional stylometric features relate to sensorial language could provide insights into human cognition and language processing. This relationship is particularly important given cognitive science theories suggesting that linguistic processes are closely tied to the brain's perceptual, motor, and introspective systems \citep{barsalou2008grounded}. Just as we know that depression can directly impact how people perceive colors \citep{bubl2010seeing}, changes in psychological states might systematically affect how people use sensory language. Thus, our goal is to investigate the relationship between these two major dimensions of linguistic style to better understand how our minds integrate sensory and psychological experiences in language use.

Our motivation for studying this relationship stems from theories in cognitive science. The interaction between different dimensions of linguistic style can be modeled using cognitive frameworks similar to the `mental lexicon' proposed by \citet{levelt1992accessing}, which posits a central repository of linguistic knowledge that mediates various aspects of language processing. 
Just as the mental lexicon provides a unified architecture for understanding how different linguistic components interact in human cognition, language models can serve as computational analogues that allow us to systematically explore the relationships between traditional stylometric features and sensorial style.

Our work aims adds to the stylometric literature by 
computationally modeling the relationship between traditional style features and sensorial style.
We introduce Stylometrically Lean Interpretable Models (SLIM-LLMs), that provide a more interpretable lens to study the relationship between traditional linguistic style and sensorial style. 
We use SLIM-LLMs to test if reduced models still benefit from LIWC features. 
In particular, we ask the following questions:
\vspace{-0.5em}
\begin{itemize}
\item \textbf{RQ 1:} When predicting sensorial language use from traditional stylistic features (like LIWC), can these style features be effectively represented in a lower-dimensional space while maintaining their predictive power for sensorial word prediction?
\end{itemize}
\vspace{-0.5em}
We model the interactions between LIWC-style and sensorial style using Reduced-Rank Ridge Regression (R4). We use R4 to identify low-rank group structures within LIWC-style.
\vspace{-0.5em}
\begin{itemize}
\item \textbf{RQ 2:} Can SLIM-LLMs models match the performance of full-scale models in predicting sensorial language use?
\end{itemize}
\vspace{-0.5em}
We conduct large-scale analysis across diverse text genres, providing empirical support for theoretical claims about the interaction between different aspects of linguistic style.

\section{Related Works}

The study of sensorial style is a relatively new area of research. There are no directly comparable studies examining sensorial style and its relation to traditional styles. Instead, we review works from allied fields ---  stylometry and sensorial linguistics --- that intersect with our work.

\subsection{Stylometry}
Stylometry focuses on analyzing linguistic style use through various computational and statistical techniques. While much of stylometric research has centered on author attribution \citep{overdorf2016blogs}, more recently stylometrics have been used to analyze emotional and psychological dimensions of language use.

One of the primary stylometric methods that focus on psycholinguistics is Linguistic Inquiry and Word Count (LIWC) \citep{pennebaker2007linguistic}. LIWC measures various linguistic features, including emotional tone, cognitive processes, and personal concerns. It has been widely used for tasks ranging from author attribution to modeling psychological states such as depression \citep{de2013predicting}.

Similarly, ANEW \citep{bradley1999affective}, provides a set of normative emotional ratings for around 1000 English words. VADER \citep{hutto2014vader} has emerged as a rule-based sentiment analysis tool that combines a lexicon and rule-based approach to measure sentiment.

In addition to these emotion-focused measures, stylometric features have traditionally included a range of measures like Readability and n-gram usage \citep{potthast2017stylometric} that represent different dimensions of linguistic style.

Recently LLMs have been increasingly utilized to represent linguistic style. \citet{li2019exploiting} and \citet{sousa2019bert} have demonstrated the effectiveness of LLMs like BERT, in modeling various aspects of linguistic style, including sentiment. However, while these LLM-based approaches have shown impressive results, they often lack interpretability. Additionally, there has been a limited focus on understanding sensorial style in these approaches.

\subsection{Sensorial Linguistics}
 Sensorial linguistics has traditionally focused on the five classical senses: visual, auditory, olfactory, gustatory, and haptic. However, recent research has expanded this model to include interoception as a sixth sense \citep{lynott2020lancaster}.

\citet{wintervision} analyzed the distribution of sensorial language across different parts of speech (nouns, adjectives, and verbs) and found that visual language dominates across all categories. This aligns with \citet{viberg1983verbs}'s proposed universal hierarchy of the senses, with vision at the top, followed by hearing, touch, smell, and taste. 

\citet{lynott2020lancaster} introduced the Sensorimotor Lexicon, a comprehensive resource containing sensory ratings for around 40,000 concepts across six sensory dimensions, including interoception.

\subsection{Sensorial Style}

Recently, methods have been proposed to analyze sensorial style. \citet{kernot2016impact} proposed a method to analyze sensorial style by measuring the use of sensory adjectives. \citet{khalid2022smells} introduced a method to measure sensorial style based on synaesthesia, or the propensity to replace one sensorial modality with another.

Prior works have focused on analyzing traditional stylometry and sensorial style independently, and there remains a gap in understanding how these two aspects of linguistic style interact. Our work aims to bridge this gap by proposing a novel approach that models the relationship between traditional linguistic style (as captured by LIWC features) and sensorial style.

\section{Methods}

\subsection{Representing Sensorial Style}

Sensorial style can be modeled and represented across a range of granularities. A synaesthesia-based approach has been used to model sensorial style at a high level \citep{khalid2022smells} that focuses on patterns of sensory language-use across broader linguistic units or entire texts, rather than on individual words. In contrast, we model sensorial style at the word-level, which focuses on individual sensorial words and their relationships to other linguistic style features.


We represent a sensorial sentence as a one-hot encoding of the sensorial vocabulary. \citet{khalid2022smells} have defined the sensorial vocabulary $V$ as a subset of 18,749 words from the Lancaster Sensorimotor Lexicon \citep{lynott2020lancaster}. They consider a sentence to be sensorial if it has one or more sensorial words in it. We use this criterion and consider a sensorial sentence to have just one sensorial term. For example, `it is a noisy room' has two sensorial words, the auditory `noisy' and the visual `room'. Assuming `noisy' and `room' are the second and fourth words in the sensorial vocabulary, this sentence constitutes two sensorial sentences represented as [0, 1, 0, 0, \ldots, 0] for `noisy' and [0, 0, 0, 1, \ldots, 0] for `room'. The length of the two vectors equals the size of our sensorial vocabulary; that is, $|V| = 18,749$.

We formalize the previous idea as follows. Let $V = \{w_1, w_2, ..., w_n\}$ be the sensorial vocabulary of size $n$. For a given sensorial word $w$ in a sentence, we represent it as a vector $\mathbf{y} \in \{0,1\}^n$, where  
$y_i = 1$ if $w = w_i$ and $0$ otherwise. A sentence $S$ with $m$ sensorial words is represented as a set of $m$ $n$-vectors and  $S = \{\mathbf{y}_1, \mathbf{y}_2, ..., \mathbf{y}_m\} $, where $\mathbf{y}_j$ $(j=1, \ldots, m)$ corresponds to the one hot encoding of the $j$th sensorial sentence.

We represent each sensorial sentence as a vector based on the LIWC-style. Let $X = \{x_1, x_2, ..., x_m\}$ be the set of $m$ LIWC categories. For a given sensorial sentence $S$, we exclude the sensorial term $w_s$ and represent the style of the remaining sentence as a vector $\mathbf{s} \in \mathbb{R}^m$. Each element $s_i$ of this vector corresponds to the proportion of words in $S$ excluding $w_s$ that belong to the $i^{th}$ LIWC category $x_i$: $s_i = (\big |\{w \in S \setminus \{w_s\} : w \in x_i\} \big|) / {(|S| - 1)}$. 

For example, given the sentence `it is a noisy room' with two sensorial words `noisy' and `room', we create two style vectors. For `room', the style vector will be based on [`it', `is', `a', `noisy'], and for `noisy' the style vector will be based on [`it', `is', `a', `room']. Given there are 4 words in the sentence, the style vectors for both sentences would have a value of $0.75$ in the function word dimension corresponding to `it', `is' and `a'.

\subsection{Linear Models for Style Interactions}


We use regression to model the relation between traditional style and sensorial style. Let the style features of a sentence $S$ be the LIWC vector $\mathbf{x} = (x_1, \ldots x_m)$ and let $\mathbf{y} = (y_1,y_2 \ldots y_n)$ be the one-hot sensorial vector of the sentence, where $m$ is the number of style features and $n$ is the size of the sensorial vocabulary $\mathcal{S}$. Then, $\mathbf{y}^\top = \mathbf{x}^\top\mathbf{B}+\mathbf{e}^\top$ models the relation between linguistic style $\mathbf{x}$ and sensorial language use $\mathbf{y}$, with $\mathbf{e}$ denoting the errors independent of $\mathbf{x}$. The regression coefficient matrix is $\mathbf{B}\in\mathbb{R}^{m\times n}$, and the element $b_{ij}$ is the mean increase in the sensorial word $y_j$ for a unit increase in style feature $x_i$, given other features in $\mathbf{x}$ remain unchanged.
The linear regression model is equivalent to a sensorial-word-prediction problem, where we predict the sensorial word $w_s$ in a sentence from the linguistic style of the remaining text. This method is analogous to the masked word prediction task used to train LLMs like BERT \cite{devlin2018bert}.

We fit the regression model to the training data as follows. For a set of $k$ sentences, the $i$th sentence has sensorial vector $\mathbf{y}_i = (y_{i1}, \ldots, y_{in})$, and its corresponding style vector is $\mathbf{x}_i = (x_{i1}, \ldots, x_{im})$. The training data are represented as the $k \times n$ matrix $\mathbf{Y} = [\mathbf{y}_1, \ldots, \mathbf{y}_k]^\top$ and $k \times m$ matrix $\mathbf{X} = [\mathbf{x}_1, \ldots, \mathbf{x}_k]^\top$. For a sufficiently large $k$, the least squares estimate of $\mathbf{B}$ is $(\mathbf{X}^\top \mathbf{X})^{-1} \mathbf{X}^\top \mathbf{Y}$ \citep{qian2022large}. Previous works have shown that LIWC features have a low-rank structure \citep{geng2020understanding}. However, the standard least squares approach fails to capture this structure and the latent dependencies between the sensorial features and LIWC-style features, which correspond to the columns of $\mathbf{Y}$ and $\mathbf{X}$. This limitation is particularly significant because not all LIWC features capture the same amount of information. For example, the function words category is more informative than categories like fillers. Additionally, LIWC categories exhibit hierarchical relationships and overlapping memberships. For instance, in the LIWC features, first person singular is a subcategory of personal pronouns, whereas the ingestion category contains words like `\textit{eat}' that also belong to the verb category.


\subsection{Reduced-Rank Ridge Regression}

We circumvent the previous limitations by assuming that $\mathbf{B}$ is a low-rank matrix. This assumption implies that the previous linear model becomes a reduced-rank regression model \citep{anderson1951estimating}, which assumes that $\mathbf{B}$ has a rank $r$ and $r \ll \text{min}\{m,n\}$. In a sparse $\mathbf{B}$,  a large fraction of the entries are 0, where $b_{ij}=0$ denotes that $x_i$ and $y_j$ are not associated. Similarly, a row sparse $\mathbf{B}$ has  $b_{ij} = 0$ for $j = 1, \ldots, n$ for many $i$s. If the $i$th row of $\mathbf{B}$ is zero, then  $x_i$ is not associated with any sensorial word. To model a rank-$r$ $\mathbf{B}$, we set  $\mathbf{B}=\mathbf{U}\mathbf{V}^\top$, where $\mathbf{U} = (u_1,u_2 \ldots u_r) \in \mathbb{R}^{m\times r}$ and $\mathbf{V} = (v_1,v_2 \ldots v_r) \in \mathbb{R}^{n\times r}$. By assuming row sparsity of $\mathbf{B}$, we can effectively select a subset of LIWC features that have the strongest associations with sensorial words across different contexts. This assumption is more appropriate for our goals of identifying the most influential LIWC features that contribute to sensorial language use.

Consider a reduced-rank model for regressing $\mathbf{Y}$ on $\mathbf{X}$. For  a rank $r$, \citet{chen2012sparse} propose a (row) sparse reduced-rank regression (SRRR) of $\mathbf{B}$ via $\mathbf{U}$ and $\mathbf{V}$ estimates as
\begin{align}\label{eq:srrr}
\hat{\mathbf{U}}_s, \hat{\mathbf{V}}_s = \argmin_{\substack{\mathbf{U} \in \mathbb{R}^{m \times r}\\ \mathbf{V}^\top \mathbf{V} = \mathbf{I}_r}} 
&\Big\{\frac{1}{2}\|\mathbf{Y} - \mathbf{XUV}^\top\|^2_F \nonumber \\
&\quad + \lambda \sum_{j=1}^m \|\mathbf{U}_j\|_2\Big\}
\end{align}
where $\hat {\mathbf{B}}_s $ is the SRRR estimate of $\mathbf{B}$, $\mathbf{I}_r$ is an $r \times r$ identity matrix, $\|\cdot \|_F$ is the Frobenius norm, and $\|\mathbf{U}_{j}\|_2$ is the group lasso penalty on the $j$th row of $\mathbf{U}$ \citep{yuan2006model}. \citet{qian2022large} develop an efficient alternative minimization algorithm for estimating $\mathbf{U}$ and $\mathbf{V}$, which estimates $\mathbf{U}$ given $\mathbf{V}$ and vice versa. The group lasso norm on $\mathbf{U}$ rows implies that some of the $\mathbf{B}_s$ rows are zeros, but the estimation algorithm suffers from computational bottlenecks particularly when $k$ and $m$ are in the order of ten thousand.

We propose Reduced-Rank Ridge Regression (R4) as an efficient alternative to SRRR. The $\mathbf{B}$ matrix in our problem is not sparse because all stylistic features are associated with sensorial words, even when their magnitudes are small; therefore, we replace the group lasso penalty on the $\mathbf{B}$ rows by a ridge penalty to obtain the R4 estimates of $\mathbf{U}$ and $\mathbf{V}$ as 
\begin{align}\label{eq:r4}
\hat{\mathbf{U}}, \hat{\mathbf{V}} = \argmin_{\substack{\mathbf{U} \in \mathbb{R}^{m \times r}\\ \mathbf{V}^\top \mathbf{V} = \mathbf{I}_r}} 
&\Big\{\frac{1}{2}\|\mathbf{Y} - \mathbf{XUV}^\top\|^2_F \nonumber \\
&\quad + \lambda \sum_{j=1}^m \|\mathbf{U}_j\|^2_2\Big\}, \nonumber \\
&\hat{\mathbf{B}} = \hat{\mathbf{U}}\hat{\mathbf{V}}^\top
\end{align}
where $\hat {\mathbf{B}}$ is the R4 estimate of $\mathbf{B}$ and is obtained by a slight modification of the alternative minimization algorithm in \citet{qian2022large}. The estimation algorithm of $\mathbf{V}$ given $\mathbf{U}$ remains the same in \eqref{eq:srrr}, but the estimation of $\mathbf{U}$ given $\mathbf{V}$ uses ridge regression. Unlike $\mathbf{B}_s$ in \eqref{eq:srrr}, $\hat{\mathbf{B}}$  is not sparse but has better predictive performance \citep{Has20}. The columns of $\hat {\mathbf{U}}$ represent the latent factors or components that capture the shared structure between LIWC and sensorial features. 

By reducing LIWC features to rank $r\ll\min\{m,n\}$, we can identify if a small set of latent dimensions captures key stylistic information.

\subsection{Modeling Non-Linear Style Interactions}\label{sec:nonlinear_model}

The R4 model in \eqref{eq:r4} assumes a linear association between LIWC-style and sensorial style. The associations, however, are not linear from linguistic and cognitive perspectives. We model the relationship between LIWC-style and sensorial style using LLMs as a proxy. LLMs, trained on vast corpora of human language, encapsulate general language norms and patterns. They capture the complex interactions mediated by our broader linguistic knowledge and cognitive processes \citep{manning2020emergent}.

To model this interaction, we represent traditional stylistic features of a sentence using our LIWC-based representation. We then use an LLM for a masked language modeling task on the original sentence, with the sensorial words masked. Finally, we use the LLM's predictions for masked sensorial words, combined with the LIWC-style, to predict sensorial style. Formally, let $S$ be the original sentence, and $m(S)$ be the sentence with sensorial words masked. Let $f$ be the function represented by the LLM that takes the masked sentence $m(S)$ and returns the encoder embedding representation of the masked word. Then, the model relating sensorial words and LLM's encoder embeddings of the masked word is 
\begin{align}\label{eq:llm}
\mathbf{y}_i &= g(f(m(S_i)); \mathbf{x}_i) + \mathbf{e}_i,  \mathbf{e}_i \in \mathbb{R}^n
\end{align}

where $S_i$ is the $i$th sentence, $\mathbf{y}_i$ and $\mathbf{x}_i$ remain the same as in \eqref{eq:r4},  $\mathbf{e}_i$ is the $i$th error  vector, and $g$ is a classifier function that predicts sensorial language use from the combination of the LLM's encoder embeddings and the original stylistic features.

\subsection{Stylometrically Lean Interpretable Models (SLIM-LLMs)}

LLMs like BERT are often overparameterized \citep{matton2019emergent} and may learn LIWC-like features implicitly. By reducing parameters, we can test if explicit LIWC features still provide complementary information, suggesting they capture fundamental style dimensions. 
We propose using dimensionality reduction techniques to create Stylometrically Lean Interpretable Models (SLIM-LLMs). SLIM-LLMs are reduced versions of standard LLMs that aim to reveal the underlying relationships between LIWC-style and sensorial style more clearly. We create SLIM-LLMs using Singular Value Decomposition (SVD). Let $\mathbf{E} \in \mathbb{R}^{k \times d}$ be the encoder embedding matrix of our LLM, where $d$ is the dimension of the hidden state and $k$ is the number of sentences in our dataset.

The SLIM-LLM retain only the top $r$ singular values and their corresponding singular vectors for the SVD of $\mathbf{E}$ and are denoted as $\mathbf{E}_{\text{slim}}$. Specifically, let $\mathbf{E} = \mathbf{U}\mathbf{\Sigma}\mathbf{V}^\top$ be the SVD of $\mathbf{E}$, where $\mathbf{U} \in \mathbb{R}^{k \times k}$ and $\mathbf{V} \in \mathbb{R}^{d \times d}$ are the left and right orthonormal matrices. Then, $\mathbf{E}_{\text{slim}} = \mathbf{U}_r \mathbf{\Sigma}_r \mathbf{V}_r^\top$, where $\mathbf{U}_r \in \mathbb{R}^{k \times r}$, $\mathbf{\Sigma}_r \in \mathbb{R}^{r \times r}$, and $\mathbf{V}_r \in \mathbb{R}^{d \times r}$. The nonlinear classification model relating sensorial words and LLMs in \eqref{eq:llm} is now rewritten for SLIM-LLMs as
\begin{align}\label{eq:sllm}
\mathbf{y}_i &= g(f_{\text{slim}}(m(S_i)); \mathbf{x}_i) + \mathbf{e}_{\text{slim}i}, \mathbf{e}_{\text{slim}i} \in \mathbb{R}^n
\end{align}
where  $\mathbf{e}_{\text{slim}i}$ is the $i$th error term, $f_{\text{slim}}$ is the function represented by our SLIM-LLM that takes the masked sentence $m(S_i)$ as input and outputs a dimension-reduced embedding of $\mathbf{x}_i$, and $g$ is a classifier function that predicts sensorial language use from the combination of the SLIM-LLM's reduced encoder embeddings and the original stylistic features. In this formulation, $f_{\text{slim}}(m(S_i))$ represents the projection of the masked sentence $m(S_i)$ onto the reduced-dimensional space defined by $\mathbf{U}_r$ so that $f_{\text{slim}}(m(S_i)) = \mathbf{U}_r^\top f(m(S_i))$,
where $f(m(S_i))$ is the original LLM's encoder embedding for the masked sentence $m(S_i)$. By reducing the dimensionality of the encoder embeddings, we aim to maintain the benefits of using LLMs as proxies for the mental lexicon while revealing more interpretable relationships between the different aspects of linguistic style.

The choice of $r$, the number of singular values to retain, represents a trade-off between model complexity and interpretability. A smaller $r$ results in a more interpretable model, but may lose some nuanced relationships, while a larger $r$ retains more information but may be less interpretable. The optimal value of $r$ can be determined through empirical analysis.

\begin{figure*}[!htbp]
 \centering
 \includegraphics[width=1.0\textwidth]{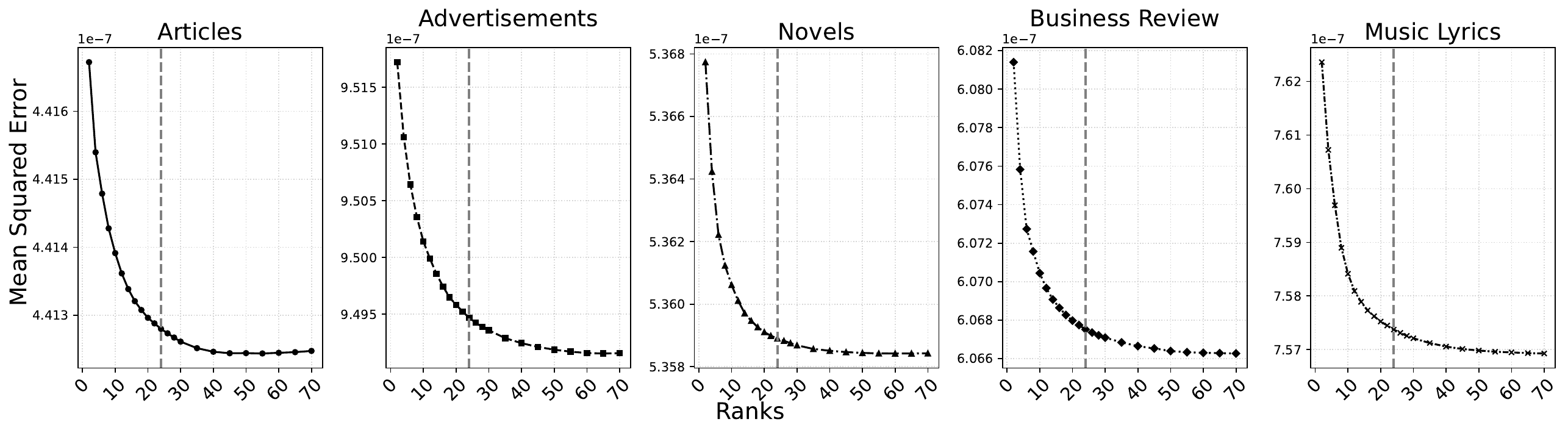} 
 \caption{Mean Squared Error (MSE) for the five language aspect datasets (Articles, Advertisements, Novels, Business Reviews, and Music Lyrics) plotted against the number of latent dimensions ($r$) in the Reduced-Rank Ridge Regression (R4) model. The plot shows the decrease in reconstruction error as the number of latent dimensions increases from 1 to 74.}
 \label{fig:mse_plot}
\end{figure*}

\section{Datasets}
We study the style of 5 different text genres using use BERT-base \citep{devlin2018bert}\footnote{Experiments using BERT-large, DistilBERT \citep{sanh2019distilbert} and RoBERTa-base \citep{liu2019roberta} gave comparable results (See: Appendix \ref{app:model_comparison}), thus we only report BERT-base results.}. This section details the datasets used in our study.

Each genre represents a distinct way in which language is employed to achieve specific communicative goals or to serve particular purposes.

\textbf{Critical Language:} Reviews from the \href{https://web.archive.org/web/20190213170507/https://www.yelp.com/dataset/challenge}{Yelp Dataset Challenge} (2005-2013), encompassing approximately 42,000 businesses.

\textbf{Literary Language:} English novels from Project Gutenberg's Domestic fiction category, spanning works from $18^{th}$ century author Regina Maria Roche to $20^{th}$ century writer Lucy Maud Montgomery.

\textbf{Poetic Language:} Lyrics of songs featured on the Billboard Hot 100 charts (1963-2021), obtained via the Genius API. This chart is widely regarded as the music industry benchmark \citep{whitburn2010billboard}.

\textbf{Persuasive Language:} Airbnb property descriptions (2008-2022), showcasing accommodations, amenities, and local attractions to potential guests.

\textbf{Informative Language:}  Wikipedia articles, collected in July 2024. Unlike other datasets, these entries are subject to continuous updates, precluding precise dating.

Table \ref{tab:text-collections} (Appendix \ref{app:table}) presents an overview of our text collections and genres, along with the specific number of sensorial sentences extracted from each collection. For our experiments, we randomly select a sample of 300,000 sensorial sentences from each set to ensure consistency across all genres.

\begin{figure*}[!htbp]
 \centering
 \includegraphics[width=1\textwidth]{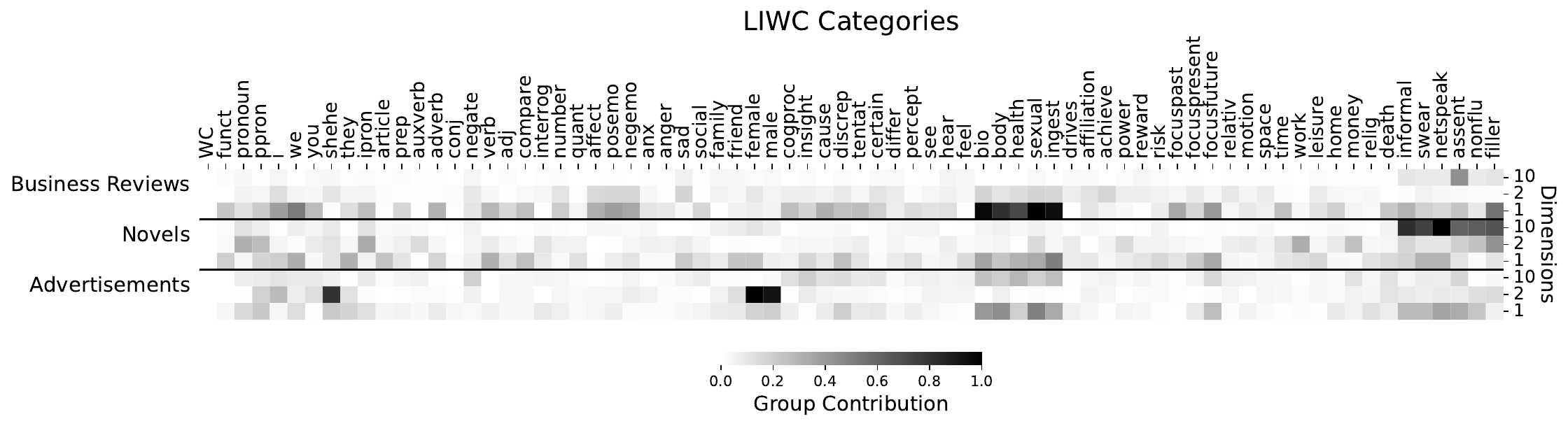} 
 
 \caption{The heatmap shows the contribution of LIWC categories to specific latent dimensions, across three genres: Business Reviews, Novels, and Advertisements.}
\label{fig:combined_latent_groups}
\end{figure*}

\section{Results}
\subsection{Latent Representation of LIWC-Style}

We investigate the relationship between the latent representation of LIWC-style and sensorial style. To find the optimal number of latent dimensions that best capture LIWC-style, we solve the Reduced-Rank Ridge Regression (R4) for a range of $r$ values from 1 to 74.

We calculate the mean squared error (MSE) of the reconstructed $\mathbf{B}=\mathbf{U}\mathbf{V}^\top$ for this range of $r$ on the test data. Figure \ref{fig:mse_plot} shows the MSE for the five datasets across different values of $r$.

While the reconstruction errors vary in absolute terms between the five genres, we observe a general trend across all datasets. On average, we see the greatest decrease in the reconstruction error within the first 20 dimensions. The error rate begins to asymptote for values of $r > 20$.

Based on this observation and the diminishing returns in error reduction, we empirically determine that $r \approx 24$ provides an optimal latent dimension representation for LIWC-style. 

\begin{figure}[b!]
    \centering
    \includegraphics[width=0.5\textwidth]{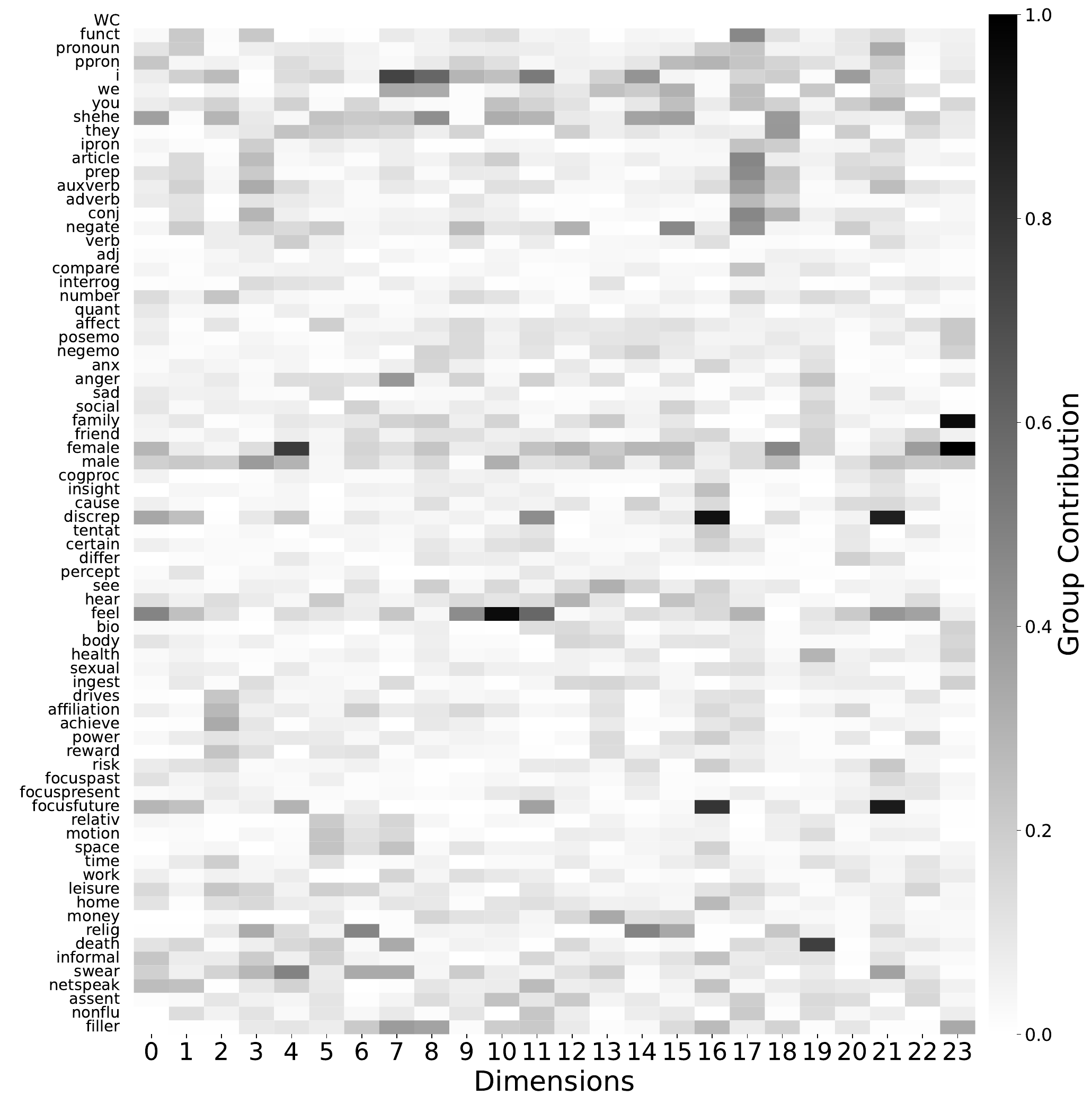}
    \caption{Heatmap showing the latent representation of LIWC categories across 24 dimensions for Wikipedia articles. The intensity indicates the strength of the contribution of each LIWC category to each latent dimension.}
    \label{fig:latent_groups.png}
\end{figure}

This finding suggests that the relationship between LIWC-style features and sensorial language use can be effectively represented in a relatively low-dimensional latent space across diverse language genres while maintaining predictive performance.

\subsection{Group structure in LIWC-Style}

In the original formulation of our model, $\mathbf{y}^\top = \mathbf{x}^\top\mathbf{B}+e^\top$, all dimensions of the LIWC features are treated as independent. However, our analysis of the $\mathbf{U} \in \mathbb{R}^{n \times r}$ matrix, which represents the latent dimensions of our R4 model, reveals group structures indicating inter-dependencies among LIWC features and their collective relationship with sensorial style.

Figure \ref{fig:latent_groups.png} illustrates the group structure in the  $\mathbf{U} \in \mathbb{R}^{74 \times 24}$ latent representation for Wikipedia articles. We find similar group structures in the latent representations of other genres as well\footnote{See Appendix \ref{app:heatmap} for the representations of \\
other genres and more detailed visualizations.}. From the figure, we note that some latent dimensions appear more influential than others, as indicated by stronger and more widespread contributions across LIWC categories, as an example the Discrepancy category \textit{`discrep'} contributes to both the $16^{th}$ and the $21^{st}$ dimensions. We also find that related LIWC categories often contribute strongly to the same latent dimensions, forming natural groupings. An example of this would be the contribution of function words, categories like \textit{`i'}, \textit{`shehe'}, \textit{`we'} (corresponding to $1^{st}$, $2^{nd}$, and $3^{rd}$ person pronouns respectively) in $17^{th}$ dimension.

In Figure \ref{fig:combined_latent_groups}, we examine a sample of columns of 3 other genres. We observe that:

\noindent \textbf{Business Reviews (Yelp):} A group forms around categories of LIWC biological processes, including words focused on consumption (dimension 1). This aligns with the nature of restaurant reviews, where descriptions of food and eating experiences are central.

\noindent \textbf{Novels (Gutenberg):} We observe a group forming around informal language use, including categories related to fillers, non-fluencies, and netspeak (dimension 10). This clustering would reflect the author's attempt to mimic natural, conversational speech patterns in dialogue and narration.

\noindent \textbf{Advertisements (Airbnb):} We observe an emergent group, not apparent in the standard LIWC classification, that combines elements from disparate LIWC categories, specifically gendered words (masculine and feminine) from the social processes category and gendered pronouns (she/he) from the function word category (dimension 2). This would suggest that Airbnb property descriptions may employ gender-specific language strategies. This finding demonstrates how our approach can reveal latent linguistic structures that are not immediately evident from simple LIWC groupings.

We find that these genre-specific group structures, emerging naturally from our latent representation analysis. These latent representations retain the predictive power, while not being constrained by the original independent dimension assumption of full LIWC.


\subsection{Exploring LIWC-Style using SLIM-BERT}

\begin{figure*}[!t]
 \centering
 \includegraphics[width=1\textwidth]{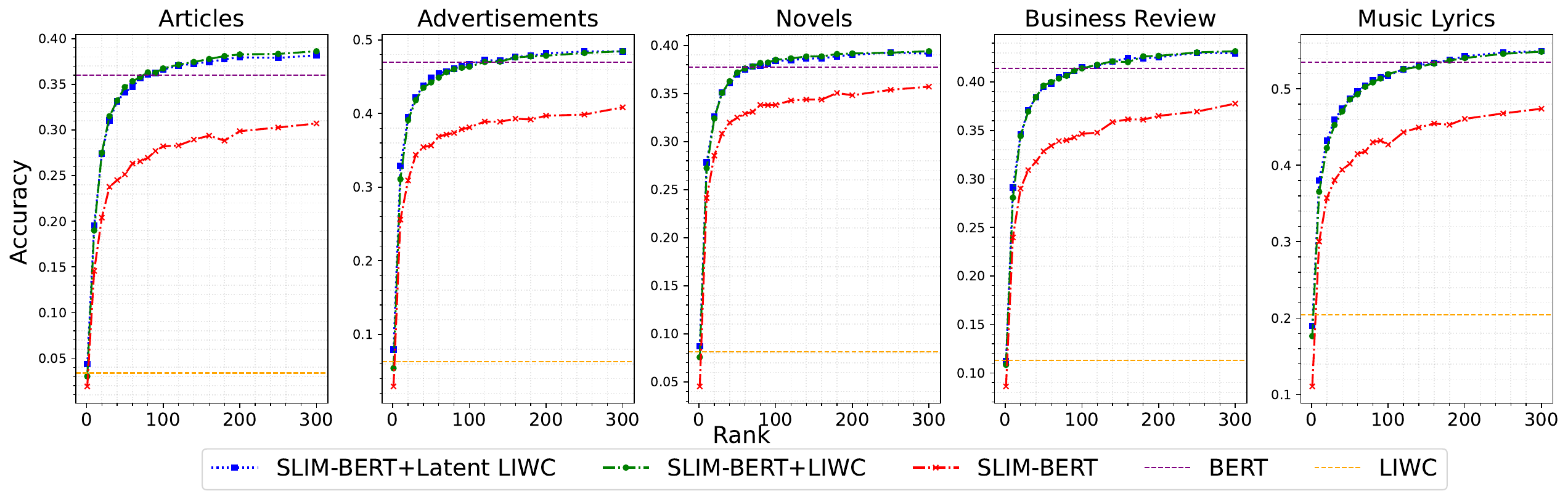} 
 
 \caption{Accuracy of sensorial word prediction against the rank (number of dimensions) used in the SLIM-BERT model for different language aspects}
\label{fig:acc_latent}
\end{figure*}

We investigate the relationship between linguistic style and sensorial language use by using low-dimensional projections of LLMs --- SLIM-LLMs model augmented with LIWC features. We use these SLIM-LLMs for the sensorial word prediction task described in section \ref{sec:nonlinear_model}. For each masked sensorial sentence, we extract the SLIM-LLM representation and use it (along with LIWC representations) as input to a fully connected Multi-Layered Perceptron (MLP) that is trained to predict the masked sensorial word. Figure \ref{fig:acc_latent} presents the performance of BERT-base for each language aspect. We focus on the first 240 dimensions of the SLIM-BERT model.

We compare the performance of three configurations of SLIM-BERT:

\noindent \textbf{SLIM-BERT+Latent LIWC}: SLIM-BERT augmented with latent LIWC features.

\noindent \textbf{SLIM-BERT+LIWC}: SLIM-BERT augmented with raw LIWC features.

\noindent \textbf{SLIM-BERT}: SLIM-BERT without LIWC features.

For reference, we also show the performance of the full BERT-base model and raw LIWC features (shown as horizontal lines).

Across all genres, we observe that augmenting SLIM-BERT with LIWC features (both latent and raw) consistently improves performance over SLIM-BERT alone. For instance, in Articles, we find that SLIM-BERT+Latent LIWC achieves an accuracy of 0.380, compared to 0.299 for SLIM-BERT alone. This pattern is consistent across other categories, with SLIM-BERT+Latent LIWC reaching accuracies of 0.483 for Advertisements, 0.390 for Novels, 0.430 for Business Reviews, and 0.545 for Music Lyrics. These results suggest that linguistic style, as captured by LIWC, provides complementary information to the language model for predicting sensorial language use.

The SLIM-BERT with Latent LIWC performs as well as or slightly better than SLIM-BERT with the raw LIWC features. For Music Lyrics, SLIM-BERT+Latent LIWC achieves 0.545 accuracy compared to 0.543 for SLIM-BERT+LIWC, indicating, the latent representation of LIWC features effectively captures the most relevant aspects of linguistic style for this task.

In most cases, our SLIM-BERT+Latent LIWC approaches or even exceeds the performance of the full BERT model, while using a fraction of the parameters. For instance, in Novels, SLIM-BERT+Latent LIWC achieves 0.390 accuracy compared to 0.378 for the full BERT model. Similarly, for Business Reviews, SLIM-BERT+Latent LIWC reaches 0.430 accuracy, surpassing the full BERT model's 0.416. This demonstrates the effectiveness of our dimensionality reduction approach in capturing the most relevant features for this task. The dimensionality reduction filters out noise and less relevant information, focusing on the most salient features of sensorial language prediction. 

These results demonstrate the effectiveness of SLIM-BERT in modeling the relationship between linguistic style and sensorial language use. The consistent improvements from LIWC augmentation, particularly using latent LIWC representation, suggest a strong link between stylometric features and sensorial language.

\section{Conclusion}

Our work demonstrates that both linguistic style (captured through LIWC features) and sensorial language use can be effectively modeled using dimensionally-reduced representations.

We found that traditional stylistic features can indeed be effectively represented in a lower-dimensional space while maintaining predictive power. Our analysis showed that a reduced latent representation with just 24 dimensions effectively captures the key stylometric information from the original 74 LIWC features across different genres of text. This dimensionality reduction not only preserves the predictive capabilities but also reveals meaningful groupings of stylistic features.

Our approach successfully demonstrated that reduced language models augmented with LIWC information can match or exceed the performance of full-scale models in predicting sensorial language use, revealing that LIWC features capture fundamental style information not learned by SLIM-LLMs.



\section{Limitations}


The main focus of this work has been on LIWC-style features. While our approach can be extended to incorporate other stylometric features such as ANEW, VADER, and measures of linguistic complexity like Readability and Hapax Legomenon, such extensions would let us not only study the relationships between these features and sensorial style, but also the interactions with the rest of the stylometric features.

Another limitation of this study is its focus on English language texts. Given that sensorial perception and its linguistic expression can vary substantially across cultures and languages, future work should explore cross-cultural applications.  The dimensionality reduction technique used to create SLIM-LLMs is not inherently language-specific and is only limited by the underlying LLM's training data. This approach can be extended to other languages by creating SLIM versions of language-specific or multilingual models, such as SLIM-BETO for Spanish (based on the BETO model \citep{CaneteCFP2020}) or SLIM-mBERT (based on the multilingual BERT model).

Like most studies in this domain, our work treats linguistic style as static, without accounting for temporal evolution. This limitation is particularly relevant when analyzing texts published by an individual over extended periods, as stylistic features and their relationship to sensorial language might shift over time.

While our model effectively captures relationships between style features and sensorial language, the current implementation of SLIM-LLMs focuses on encoder-only transformer architectures. The applicability of this approach to decoder-only architectures like GPT remains an open question for future research.

Our evaluation relies primarily on quantitative metrics of prediction accuracy. Future work could benefit from incorporating qualitative validation approaches, including native speaker judgments of sensorial language use and style relationships.

These limitations suggest several promising directions for future research, including cross-lingual studies of sensorial style, investigation of temporal dynamics in linguistic style, and extension to other model architectures. Addressing these aspects would contribute to a more complete understanding of how different dimensions of linguistic style interact with sensorial language use.


\bibliography{custom}

\begin{thebibliography}{35}
\providecommand{\natexlab}[1]{#1}

\bibitem[{Anderson(1951)}]{anderson1951estimating}
Theodore~Wilbur Anderson. 1951.
\newblock Estimating linear restrictions on regression coefficients for multivariate normal distributions.
\newblock \emph{The Annals of Mathematical Statistics}, pages 327--351.

\bibitem[{Barsalou(2008)}]{barsalou2008grounded}
Lawrence~W Barsalou. 2008.
\newblock Grounded cognition.
\newblock \emph{Annu. Rev. Psychol.}, 59(1):617--645.

\bibitem[{Beauchamp et~al.(2008)}]{beauchamp2008belmont}
Tom~L Beauchamp et~al. 2008.
\newblock The belmont report.
\newblock \emph{The Oxford textbook of clinical research ethics}, pages 149--155.

\bibitem[{Bradley and Lang(1999)}]{bradley1999affective}
Margaret~M Bradley and Peter~J Lang. 1999.
\newblock Affective norms for english words (anew): Instruction manual and affective ratings.
\newblock Technical report, Technical report C-1, the center for research in psychophysiology~….

\bibitem[{Bubl et~al.(2010)Bubl, Kern, Ebert, Bach, and Van~Elst}]{bubl2010seeing}
Emanuel Bubl, Elena Kern, Dieter Ebert, Michael Bach, and Ludger~Tebartz Van~Elst. 2010.
\newblock Seeing gray when feeling blue? depression can be measured in the eye of the diseased.
\newblock \emph{Biological psychiatry}, 68(2):205--208.

\bibitem[{Cañete et~al.(2020)Cañete, Chaperon, Fuentes, Ho, Kang, and Pérez}]{CaneteCFP2020}
José Cañete, Gabriel Chaperon, Rodrigo Fuentes, Jou-Hui Ho, Hojin Kang, and Jorge Pérez. 2020.
\newblock Spanish pre-trained bert model and evaluation data.
\newblock In \emph{PML4DC at ICLR 2020}.

\bibitem[{Chen and Huang(2012)}]{chen2012sparse}
Lisha Chen and Jianhua~Z Huang. 2012.
\newblock Sparse reduced-rank regression for simultaneous dimension reduction and variable selection.
\newblock \emph{Journal of the American Statistical Association}, 107(500):1533--1545.

\bibitem[{De~Choudhury et~al.(2013)De~Choudhury, Gamon, Counts, and Horvitz}]{de2013predicting}
Munmun De~Choudhury, Michael Gamon, Scott Counts, and Eric Horvitz. 2013.
\newblock Predicting depression via social media.
\newblock In \emph{Proceedings of the international AAAI conference on web and social media}, volume~7, pages 128--137.

\bibitem[{Devlin(2018)}]{devlin2018bert}
Jacob Devlin. 2018.
\newblock Bert: Pre-training of deep bidirectional transformers for language understanding.
\newblock \emph{arXiv preprint arXiv:1810.04805}.

\bibitem[{Geng et~al.(2020)Geng, Niu, Feng, and Huang}]{geng2020understanding}
Shuang Geng, Ben Niu, Yuanyue Feng, and Miaojia Huang. 2020.
\newblock Understanding the focal points and sentiment of learners in mooc reviews: A machine learning and sc-liwc-based approach.
\newblock \emph{British Journal of Educational Technology}, 51(5):1785--1803.

\bibitem[{Hastie(2020)}]{Has20}
Trevor Hastie. 2020.
\newblock Ridge regularization: An essential concept in data science.
\newblock \emph{Technometrics}, 62(4):426--433.

\bibitem[{{HiQ Labs, Inc. v. LinkedIn Corp.}(2019)}]{hiq_linkedin_2019}
{HiQ Labs, Inc. v. LinkedIn Corp.} 2019.
\newblock Hiq labs, inc. v. linkedin corp.

\bibitem[{Hutto and Gilbert(2014)}]{hutto2014vader}
Clayton Hutto and Eric Gilbert. 2014.
\newblock Vader: A parsimonious rule-based model for sentiment analysis of social media text.
\newblock In \emph{Proceedings of the international AAAI conference on web and social media}, volume~8, pages 216--225.

\bibitem[{Kernot et~al.(2016)Kernot, Bossomaier, and Bradbury}]{kernot2016impact}
David Kernot, Terry Bossomaier, and Roger Bradbury. 2016.
\newblock The impact of depression and apathy on sensory language.
\newblock \emph{Open Journal of Modern Linguistics}, 7(1):8--32.

\bibitem[{Khalid and Srinivasan(2022)}]{khalid2022smells}
Osama Khalid and Padmini Srinivasan. 2022.
\newblock Smells like teen spirit: An exploration of sensorial style in literary genres.
\newblock In \emph{Proceedings of the 29th International Conference on Computational Linguistics}, pages 55--64.

\bibitem[{Levelt(1992)}]{levelt1992accessing}
Willem~JM Levelt. 1992.
\newblock Accessing words in speech production: Stages, processes and representations.
\newblock \emph{Cognition}, 42(1-3):1--22.

\bibitem[{Li et~al.(2019)Li, Bing, Zhang, and Lam}]{li2019exploiting}
Xin Li, Lidong Bing, Wenxuan Zhang, and Wai Lam. 2019.
\newblock Exploiting bert for end-to-end aspect-based sentiment analysis.
\newblock In \emph{Proceedings of the 5th Workshop on Noisy User-generated Text (W-NUT 2019)}, pages 34--41.

\bibitem[{Liu(2019)}]{liu2019roberta}
Yinhan Liu. 2019.
\newblock Roberta: A robustly optimized bert pretraining approach.
\newblock \emph{arXiv preprint arXiv:1907.11692}.

\bibitem[{Lynott et~al.(2020)Lynott, Connell, Brysbaert, Brand, and Carney}]{lynott2020lancaster}
Dermot Lynott, Louise Connell, Marc Brysbaert, James Brand, and James Carney. 2020.
\newblock The lancaster sensorimotor norms: multidimensional measures of perceptual and action strength for 40,000 english words.
\newblock \emph{Behavior research methods}, 52:1271--1291.

\bibitem[{Manning et~al.(2020)Manning, Clark, Hewitt, Khandelwal, and Levy}]{manning2020emergent}
Christopher~D Manning, Kevin Clark, John Hewitt, Urvashi Khandelwal, and Omer Levy. 2020.
\newblock Emergent linguistic structure in artificial neural networks trained by self-supervision.
\newblock \emph{Proceedings of the National Academy of Sciences}, 117(48):30046--30054.

\bibitem[{Matton and de~Oliveira(2019)}]{matton2019emergent}
Alexandre Matton and Luke de~Oliveira. 2019.
\newblock Emergent properties of finetuned language representation models.
\newblock \emph{arXiv preprint arXiv:1910.10832}.

\bibitem[{Overdorf and Greenstadt(2016)}]{overdorf2016blogs}
Rebekah Overdorf and Rachel Greenstadt. 2016.
\newblock Blogs, twitter feeds, and reddit comments: Cross-domain authorship attribution.
\newblock \emph{Proceedings on Privacy Enhancing Technologies}.

\bibitem[{Pennebaker et~al.(2007)Pennebaker, Booth, and Francis}]{pennebaker2007linguistic}
James~W Pennebaker, Roger~J Booth, and Martha~E Francis. 2007.
\newblock Linguistic inquiry and word count: Liwc [computer software].
\newblock \emph{Austin, TX: liwc. net}, 135.

\bibitem[{Pennebaker et~al.(2015)Pennebaker, Boyd, Jordan, and Blackburn}]{pennebaker2015development}
James~W Pennebaker, Ryan~L Boyd, Kayla Jordan, and Kate Blackburn. 2015.
\newblock The development and psychometric properties of liwc2015.

\bibitem[{Potthast et~al.(2017)Potthast, Kiesel, Reinartz, Bevendorff, and Stein}]{potthast2017stylometric}
Martin Potthast, Johannes Kiesel, Kevin Reinartz, Janek Bevendorff, and Benno Stein. 2017.
\newblock A stylometric inquiry into hyperpartisan and fake news.
\newblock \emph{arXiv preprint arXiv:1702.05638}.

\bibitem[{Qian et~al.(2022)Qian, Tanigawa, Li, Tibshirani, Rivas, and Hastie}]{qian2022large}
Junyang Qian, Yosuke Tanigawa, Ruilin Li, Robert Tibshirani, Manuel~A Rivas, and Trevor Hastie. 2022.
\newblock Large-scale multivariate sparse regression with applications to uk biobank.
\newblock \emph{The annals of applied statistics}, 16(3):1891.

\bibitem[{Sandvig et~al.(2014)Sandvig, Hamilton, Karahalios, and Langbort}]{sandvig2014auditing}
Christian Sandvig, Kevin Hamilton, Karrie Karahalios, and Cedric Langbort. 2014.
\newblock Auditing algorithms: Research methods for detecting discrimination on internet platforms.
\newblock \emph{Data and discrimination: converting critical concerns into productive inquiry}, 22(2014):4349--4357.

\bibitem[{Sanh et~al.(2019)Sanh, Debut, Chaumond, and Wolf}]{sanh2019distilbert}
Victor Sanh, L~Debut, J~Chaumond, and T~Wolf. 2019.
\newblock Distilbert, a distilled version of bert: Smaller, faster, cheaper and lighter. arxiv 2019.
\newblock \emph{arXiv preprint arXiv:1910.01108}.

\bibitem[{Sousa et~al.(2019)Sousa, Sakiyama, de~Souza~Rodrigues, Moraes, Fernandes, and Matsubara}]{sousa2019bert}
Matheus~Gomes Sousa, Kenzo Sakiyama, Lucas de~Souza~Rodrigues, Pedro~Henrique Moraes, Eraldo~Rezende Fernandes, and Edson~Takashi Matsubara. 2019.
\newblock Bert for stock market sentiment analysis.
\newblock In \emph{2019 IEEE 31st international conference on tools with artificial intelligence (ICTAI)}, pages 1597--1601. IEEE.

\bibitem[{{Van Buren v. United States}(2021)}]{van_buren_2021}
{Van Buren v. United States}. 2021.
\newblock Van buren v. united states.

\bibitem[{Viberg(1983)}]{viberg1983verbs}
{\AA}ke Viberg. 1983.
\newblock The verbs of perception: A typological study.

\bibitem[{Whitburn(2010)}]{whitburn2010billboard}
Joel Whitburn. 2010.
\newblock \emph{The Billboard Book of Top 40 Hits: Complete Chart Information about America's Most Popular Songs and Artists, 1955-2009}.
\newblock Billboard Books.

\bibitem[{Winter(2016)}]{bodosensory}
Bodo Winter. 2016.
\newblock \emph{The sensory structure of the English lexicon}.
\newblock University of California, Merced.

\bibitem[{Winter et~al.(2018)Winter, Perlman, and Majid}]{wintervision}
Bodo Winter, Marcus Perlman, and Asifa Majid. 2018.
\newblock Vision dominates in perceptual language: English sensory vocabulary is optimized for usage.
\newblock \emph{Cognition}, 179:213--220.

\bibitem[{Yuan and Lin(2006)}]{yuan2006model}
Ming Yuan and Yi~Lin. 2006.
\newblock Model selection and estimation in regression with grouped variables.
\newblock \emph{Journal of the Royal Statistical Society Series B: Statistical Methodology}, 68(1):49--67.

\end{thebibliography}
\onecolumn

\appendix
\newpage
\section{Appendix}
\subsection{Summary of the text collections}\label{app:table}

\begin{table*}[h]
\centering
\begin{tabular}{|l|l|l|l|}
\hline
Language Genre & Datasets & Source & Sensorial Sentences \\
\hline
Critical & Business Reviews & Yelp.com &2,101,603\\
Literary & Novels & Project Gutenberg &  1,929,260\\
Poetic & Music Lyrics & Genius.com &1,107,749 \\
Persuasive & Advertisements & Airbnb Descriptions &1,442,050 \\
Informative & Articles & Wikipedia &1,563,888 \\
\hline
\end{tabular}
\caption{Overview of text collections and genres}
\label{tab:text-collections}
\end{table*}

\clearpage  

\subsection{Latent Representations of LIWC-Style Across Text Genres}\label{app:heatmap}
\begin{figure}[!b]
    \centering
    \subfloat[Music Lyrics]{
        \includegraphics[width=0.45\textwidth]{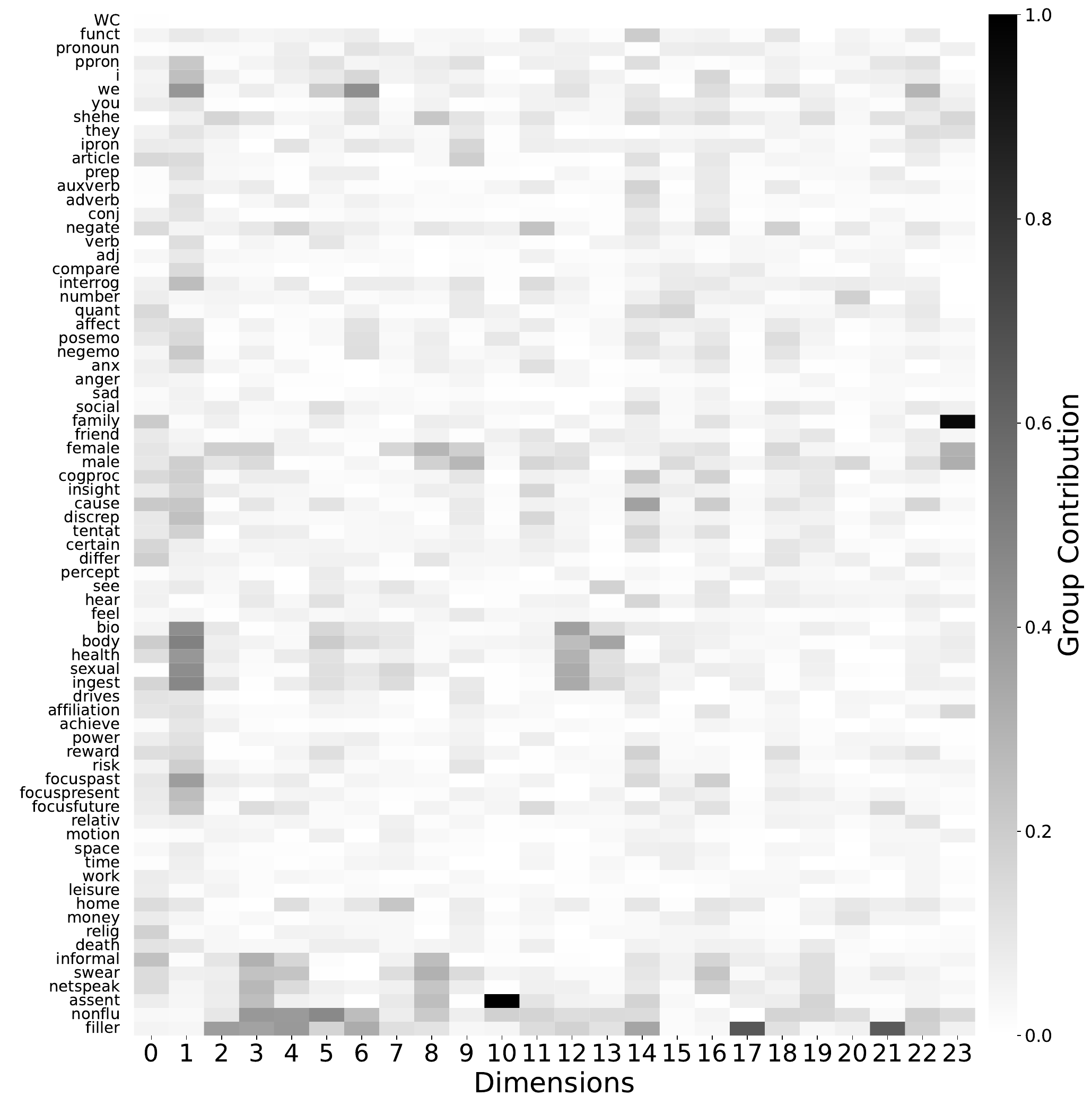}
    }
    \hfill
    \subfloat[Novels]{
        \includegraphics[width=0.45\textwidth]{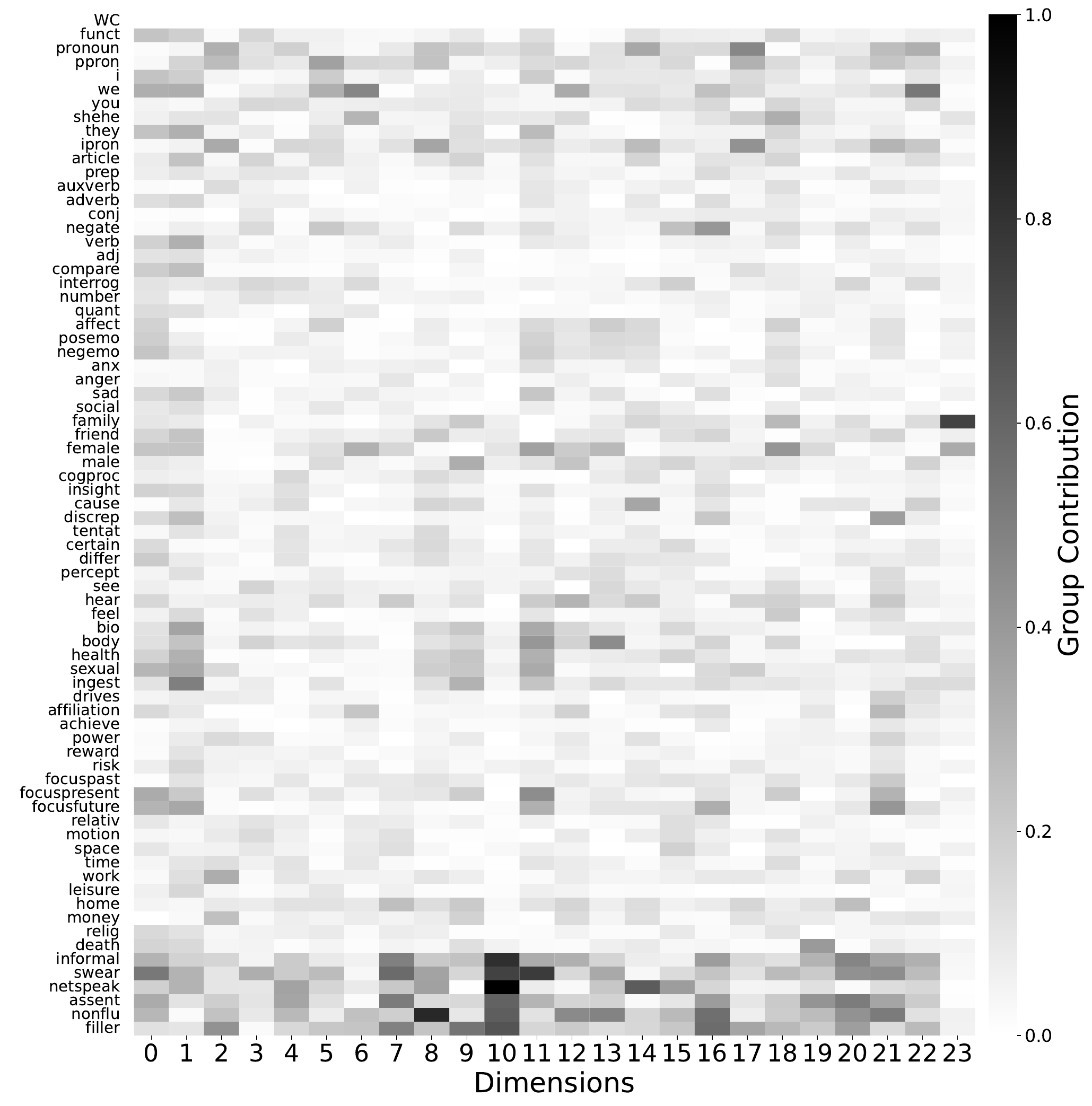}
    }
    
    \vspace{1em}
    
    \subfloat[Advertisements]{
        \includegraphics[width=0.45\textwidth]{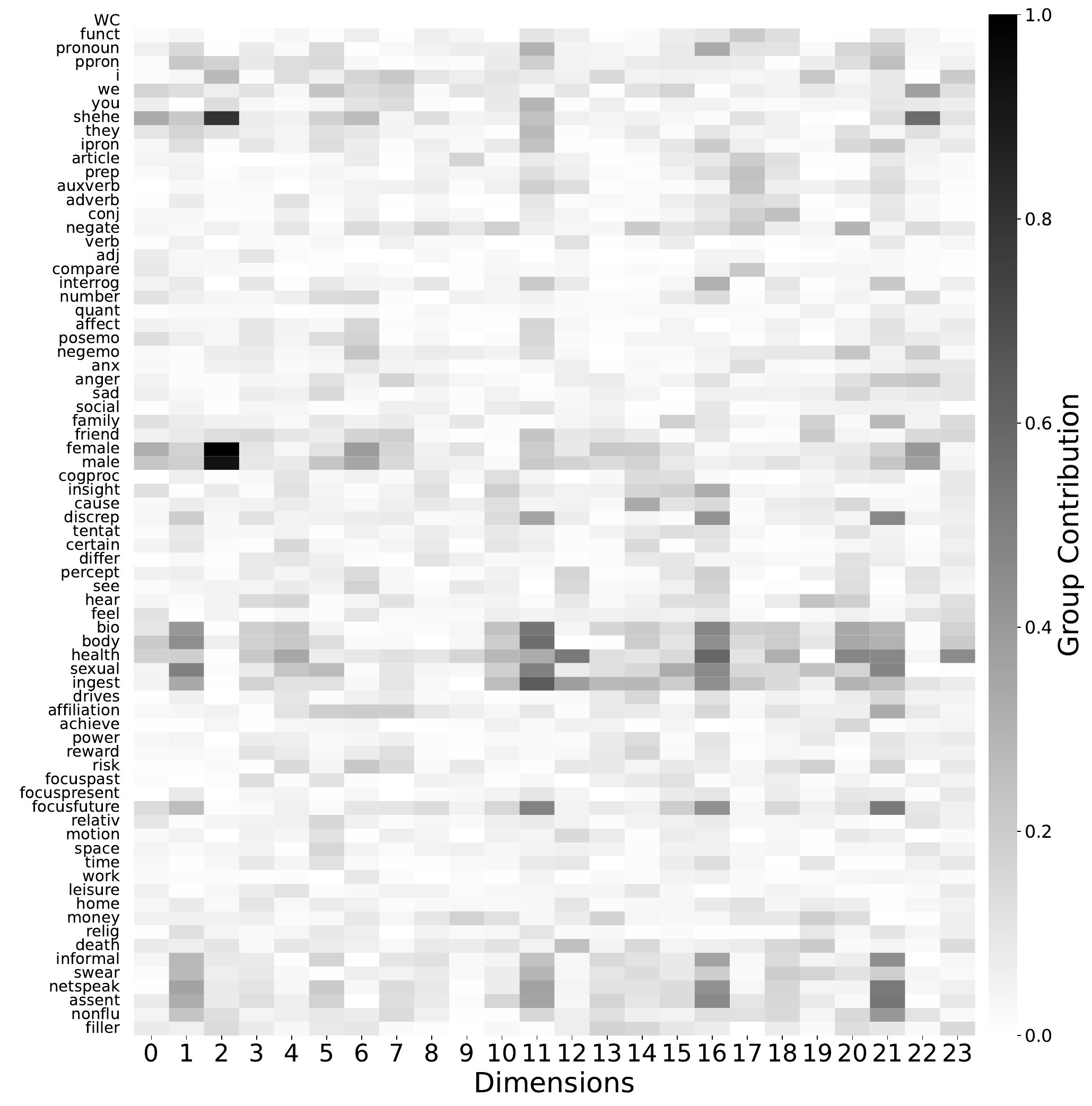}
    }
    \hfill
    \subfloat[Business Reviews]{
        \includegraphics[width=0.45\textwidth]{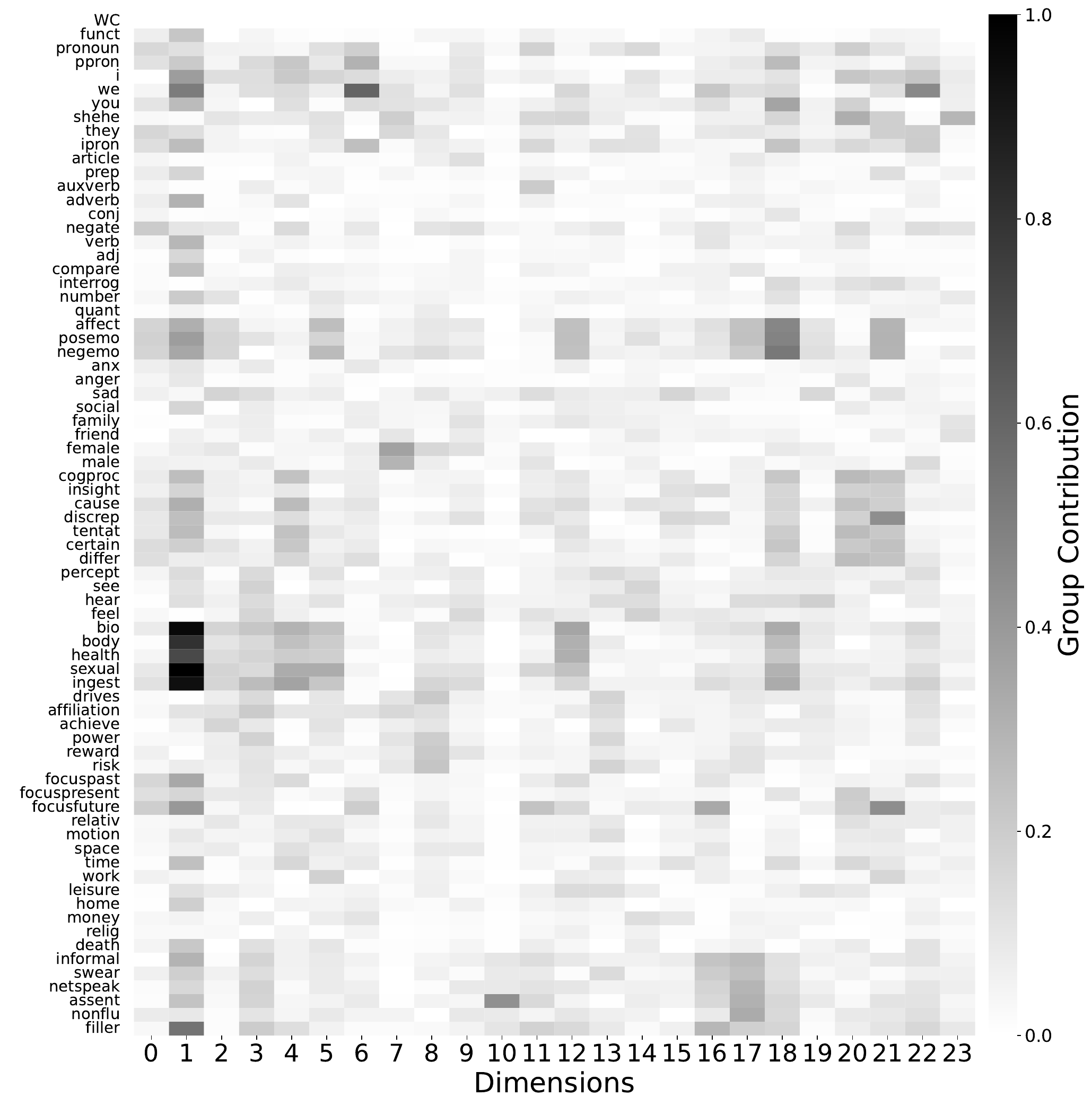}
    }
    \caption{Heatmaps showing the latent representation of LIWC categories across 24 dimensions for different text genres: (a) Music Lyrics, (b) Novels, (c) Advertisements, and (d) Business Reviews. The intensity indicates the strength of contribution of each LIWC category to each latent dimension.}
\end{figure}
\begin{figure}[!b]
    \centering
    \subfloat[Articles]{
        \includegraphics[width=0.45\textwidth]{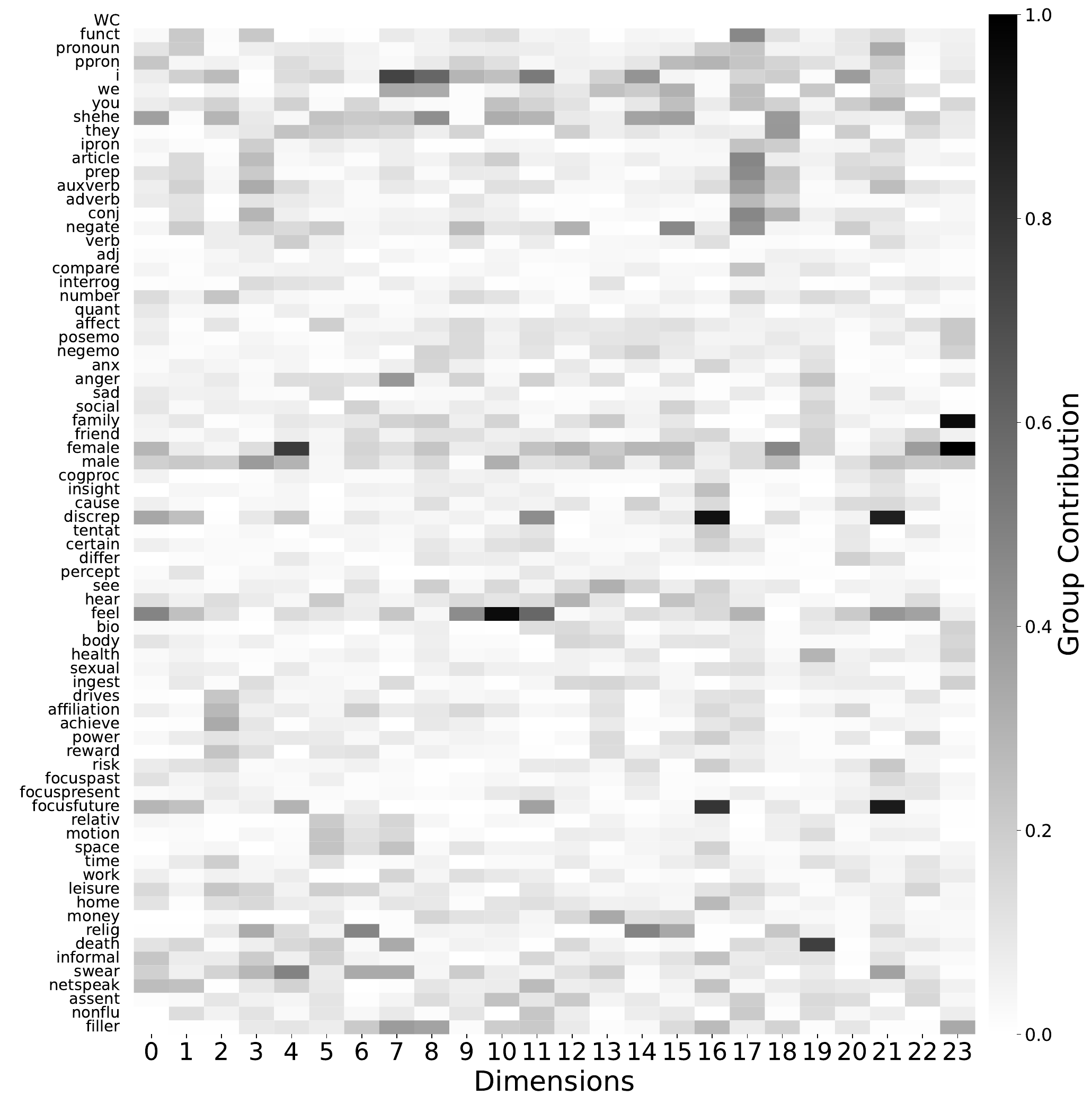}
    }\\
    \caption{Heatmap showing the latent representation of LIWC categories across 24 dimensions for (e) Articles.}
\end{figure}

\clearpage  

\begin{sidewaystable}
\centering
\label{tab:performance-comparison}

\subsection{Performance Comparison of SLIM-LLMs Across Different Text Genres}\label{app:model_comparison}
\tiny
\centering
\label{tab:performance-comparison}
\tiny
\begin{tabular}{llccccccccccccccc}
\toprule
\multirow{2}{*}{Model} & \multirow{2}{*}{Configuration} & \multicolumn{3}{c}{Articles} & \multicolumn{3}{c}{Advertisements} & \multicolumn{3}{c}{Novels} & \multicolumn{3}{c}{Business Reviews} & \multicolumn{3}{c}{Music Lyrics} \\
\cmidrule(lr){3-5} \cmidrule(lr){6-8} \cmidrule(lr){9-11} \cmidrule(lr){12-14} \cmidrule(lr){15-17}
& & SLIM-80 & SLIM-240 & Full & SLIM-80 & SLIM-240 & Full & SLIM-80 & SLIM-240 & Full & SLIM-80 & SLIM-240 & Full & SLIM-80 & SLIM-240 & Full \\
\midrule
\multirow{4}{*}{BERT-base} & Full Model & -- & -- & 0.360 & -- & -- & 0.469 & -- & -- & 0.378 & -- & -- & 0.416 & -- & -- & 0.533 \\
& SLIM & 0.279 & 0.299 & -- & 0.372 & 0.403 & -- & 0.332 & 0.352 & -- & 0.339 & 0.368 & -- & 0.423 & 0.465 & -- \\
& SLIM + LIWC & 0.362 & \textbf{0.385} & -- & 0.457 & 0.481 & -- & 0.381 & \textbf{0.391} & -- & 0.407 & 0.429 & -- & 0.511 & 0.543 & -- \\
& SLIM + Latent LIWC & 0.357 & 0.380 & -- & 0.462 & \textbf{0.483} & -- & 0.379 & 0.390 & -- & 0.409 & \textbf{0.430} & -- & 0.510 & \textbf{0.545} & -- \\
\midrule
\multirow{4}{*}{BERT-large} & Full Model & -- & -- & 0.373 & -- & -- & 0.473 & -- & -- & 0.389 & -- & -- & 0.420 & -- & -- & 0.517 \\
& SLIM & 0.289 & 0.315 & -- & 0.380 & 0.418 & -- & 0.348 & 0.364 & -- & 0.355 & 0.380 & -- & 0.434 & 0.464 & -- \\
& SLIM + LIWC & 0.384 & \textbf{0.404} & -- & 0.469 & \textbf{0.491} & -- & 0.394 & \textbf{0.406} & -- & 0.424 & \textbf{0.440} & -- & 0.513 & 0.540 & -- \\
& SLIM + Latent LIWC & 0.379 & 0.400 & -- & 0.473 & \textbf{0.491} & -- & 0.393 & 0.405 & -- & 0.424 & 0.439 & -- & 0.514 & \textbf{0.542} & -- \\
\midrule
\multirow{4}{*}{RoBERTa} & Full Model & -- & -- & 0.356 & -- & -- & 0.499 & -- & -- & 0.397 & -- & -- & 0.465 & -- & -- & 0.565 \\
& SLIM & 0.242 & 0.281 & -- & 0.386 & 0.418 & -- & 0.327 & 0.356 & -- & 0.367 & 0.401 & -- & 0.438 & 0.489 & -- \\
& SLIM + LIWC & 0.336 & \textbf{0.365} & -- & 0.472 & 0.501 & -- & 0.386 & \textbf{0.405} & -- & 0.440 & \textbf{0.467} & -- & 0.525 & 0.566 & -- \\
& SLIM + Latent LIWC & 0.336 & 0.363 & -- & 0.478 & \textbf{0.502} & -- & 0.387 & 0.403 & -- & 0.441 & 0.466 & -- & 0.528 & \textbf{0.567} & -- \\
\midrule
\multirow{4}{*}{DistilBERT} & Full Model & -- & -- & 0.330 & -- & -- & 0.454 & -- & -- & 0.348 & -- & -- & 0.391 & -- & -- & 0.523 \\
& SLIM & 0.237 & 0.267 & -- & 0.345 & 0.378 & -- & 0.290 & 0.318 & -- & 0.305 & 0.332 & -- & 0.397 & 0.446 & -- \\
& SLIM + LIWC & 0.326 & 0.347 & -- & 0.442 & \textbf{0.467} & -- & 0.343 & 0.359 & -- & 0.376 & \textbf{0.407} & -- & 0.493 & 0.532 & -- \\
& SLIM + Latent LIWC & 0.327 & \textbf{0.350} & -- & 0.441 & \textbf{0.467} & -- & 0.344 & \textbf{0.361} & -- & 0.378 & 0.400 & -- & 0.494 & \textbf{0.533} & -- \\
\midrule
LIWC & -- & -- & -- & 0.033 & -- & -- & 0.063 & -- & -- & 0.083 & -- & -- & 0.113 & -- & -- & 0.203 \\
\bottomrule
\end{tabular}
\caption*{Note: SLIM-80 and SLIM-240 refer to SLIM-LLM models with 80 and 240 dimensions respectively. \\The best performing SLIM-LLM configuration for each model and genre is highlighted in bold.}
\end{sidewaystable}

\clearpage  
\subsection{Ethical Statement}

This work aims to advance our comprehension of language patterns and stylistic relationships. 
From a risk perspective, our research approach minimizes potential negative impacts in several ways. Since we focus on analytical modeling rather than developing deployable systems, there are no direct risks associated with implementation or user-facing applications. The research design deliberately emphasizes theoretical understanding over practical application, reducing the potential for immediate misuse.

Our analysis relies entirely on public datasets accessed with appropriate permissions, and we neither collect nor process sensitive personal information. The research design explicitly avoids using demographic data or protected-class information, eliminating risks of individual re-identification or discriminatory applications.

We acknowledge, however, that any research advancing language understanding could potentially enable more sophisticated analysis tools in the future. These might include enhanced text analysis capabilities, more accurate authorship attribution, or style transfer applications. While these potential developments require significant expertise to implement, we recognize our responsibility to address these possibilities transparently. By maintaining full methodological transparency, we enable community oversight and ongoing ethical discussion.

Our risk mitigation strategy centers on three key approaches. First, we maintain complete transparency in our methods, limitations, and data processing. Second, we focus strictly on linguistic patterns rather than individual identification or demographic prediction. Third, we actively engage with ethical considerations through clear documentation and open discussion of potential applications and implications. Overall, we respect the principle of beneficence as outlined by the Belmont report \cite{beauchamp2008belmont}.

\subsection{Data Sources and Privacy}
All data used in this research is publicly available. The Yelp Dataset is a pre-anonymized public dataset. We used official APIs to collect data from Wikipedia and Genius. For Airbnb and Project Gutenberg we developed custom crawlers following ethical web scraping practices. 
All personally identifiable information (PII) was removed, including but not limited to age, gender, and demographic information. The datasets used in this work consists exclusively of English language content

\subsection{Data Collection Ethics}
Our crawls were consistent with typical auditing studies \cite{sandvig2014auditing} and are legally permissible (\citeauthor{hiq_linkedin_2019},and \citeauthor{van_buren_2021}). Data collection was conducted in compliance with each platform's Terms of Service. We adhered to ethical web scraping practices to ensure that our data collection did not interfere with user experience or the platform’s operations.

\subsection{Data Processing}

Text preprocessing included lowercasing and punctuation removal for both LIWC prediction and masked language model prediction tasks. Consistent preprocessing was applied across all datasets.

\subsection{Model Development}
We utilized embeddings from pre-trained models available on Huggingface, used in accordance with their licenses. No models were trained from scratch. Fine-tuning involved training a fully connected Multi-Layer Perceptron (MLP) on top of pre-trained embeddings. In each experiment we trained the model for 10 epochs on a Single T4 GPU. In addition, we used LIWC to extract stylometric features in accordance with its license. The results reported for language models (as shown in Figure \ref{fig:acc_latent}) represent averages from 5-fold cross-validation using a sample size of 300,000 examples.

\end{document}